\documentclass[runningheads]{llncs}
\usepackage[T1]{fontenc}
\usepackage{graphicx}
\usepackage{amsmath}
\usepackage{amssymb}
\usepackage{booktabs}
\usepackage{listings}
\usepackage{multirow}
\begin{document}
\title{GLARE: A Natural Language Interface for Querying Global Explanations
}
%
%
\author{Bhavan Vasu\thanks{Equal contribution.}\orcidID{0000-0003-4961-580X} \and \\
Rajesh Mangannavar\protect\footnotemark[1]\orcidID{0000-0001-8433-3096}}

\authorrunning{B. Vasu \and R.Mangannavar.}

\institute{Oregon State University, Corvallis OR 97330, USA 
\email{\{vasub,mangannr\}@oregonstate.edu}}

\maketitle              
\begin{abstract}

While global explanations are crucial for understanding vision models across datasets, classes, and decision contexts, their complex and monolithic nature often hinders practical exploration. Because users typically seek targeted answers to specific questions rather than static artifacts, we present an LLM-based interactive interface that provides natural language access to global explanations for black-box image classifiers. The system's core LLM acts as a mediator, translating natural language questions into structured SQL queries over local explanation data. This enables flexible aggregation without exposing users to low-level representations. For each query, the interface outputs statistics-augmented natural language responses, supporting local explanations, and intent-aligned visualizations. We evaluate the system on intent interpretation, query mapping accuracy, generalization to novel queries and datasets, and robustness to linguistic errors. Our results demonstrate that LLM-mediated querying substantially improves the accessibility and usability of global explanations for human-centered XAI.

\end{abstract}

\section{Introduction}
\label{sec:introduction}

Deep vision models have achieved remarkable success in tasks ranging from medical diagnosis to autonomous driving, yet their decision-making processes remain largely opaque. In high-stakes settings, users require explanations not only to justify individual predictions but to build a mental model of the system's behavior, assess reliability, and diagnose failure modes. Explainable AI (XAI) methods have traditionally been dichotomized into \emph{local} and \emph{global} approaches. Local methods such as saliency maps~\cite{selvaraju2017grad}, concept bottlenecks~\cite{koh2020concept}, or counterfactuals, explain specific instances (e.g., ``Why was \emph{this} image classified as a wolf?''). While useful for auditing single errors, they fail to reveal systemic biases or general reasoning patterns. Conversely, \emph{global} explanations aim to summarize the model's behavior across the entire input space, often by identifying globally important features~\cite{darwiche2022computation}~\cite{azzolin2023global} or distilling the model into a transparent surrogate such as a decision tree~\cite{lakkaraju2016interpretable}~\cite{craven1995extracting} or a Disjunctive Normal Form (DNF) formula~\cite{vasu2026local}~\cite{baugh2025neural}.

However, a critical usability gap plagues modern global explanations: they are often overwhelming in scale and complexity. For a deep network trained on a complex visual domain, a high-fidelity global explanation might consist of thousands of logical rules or prototypes. Presenting this ``explanation dump'' to a user induces cognitive overload, obscuring the very insights it aims to reveal. We argue that users rarely need a static, monolithic summary of the entire model. Instead, human explanation-seeking is an iterative, query-driven process~\cite{miller2019explanation} ~\cite{liao2020questioning}. Users approach the model with specific hypotheses or information needs, such as: \emph{``What features are necessary for the 'bedroom' class?''}, \emph{``Does the model rely on background snow to classify wolves?''}, or \emph{``Show me examples where the model relies on shape rather than texture.''} Current XAI tools force users to manually filter and aggregate local explanations to answer these questions, creating a friction that limits the practical utility of global insights.

In this paper, we present \textbf{GLARE} (Global Language-based Analysis and Retrieval of Explanations), an interactive interface that mediates between users and large-scale global explanations. Rather than treating a global explanation as a static artifact to be viewed, we treat it as a \emph{database} to be queried. We choose logical explanations ~\cite{vasu2026local}~\cite{vasu2023global}, that aggregates local Minimal Sufficient Explanations (MSXs) into a global DNF structure due to its binary nature of a concept being important vs not in the form of logical rules. We ingest these logic-based local explanations into a relational database, enabling precise structural queries over the model's reasoning patterns. Although our experiments are limited to explanations generated by ~\cite{vasu2026local}, the interface presented in the paper can work with any concept-based local explanation method.  GLARE allows users to interrogate this database using natural language. The core of our system is a Large Language Model (LLM) fine-tuned to act as a semantic parser, translating user questions into structured SQL queries. Unlike general text-to-SQL approaches, we constrain the LLM to select from a taxonomy of analytical query templates specialized in explaining decisions. Templates ranging from simple object frequency counts to complex counterfactual set operations. 
By employing a loss-masking technique during fine-tuning that focuses learning exclusively on the explanation-specific SQL structure (``fence masking''), we encourage the model to learn the \emph{relational algebra} of explanation querying rather than memorizing dataset-specific entity names, leading to zero-shot transfer to new datasets.

We evaluate GLARE on global explanations derived from the ADE20K scene parsing dataset. Our results demonstrate that the system achieves over 95\% accuracy on in-distribution queries and exhibits strong robustness to spelling errors, grammatical noise, and phrasing variations. Most notably, we demonstrate zero-shot cross-dataset transfer: a model trained exclusively on ADE20K metadata effectively interprets queries for a Pascal VOC database, a domain with a completely disjoint object vocabulary, suggesting that our approach learns generalized reasoning patterns applicable to diverse vision tasks. To summarize, Our contributions are threefold: (i) we introduce a natural-language interface for interrogating global explanations as queryable databases; (ii) we define a SQL-based intermediate representation for aggregating, filtering, and contrasting local explanations; and (iii) we demonstrate that synthetic-data fine-tuning with SQL-fence loss masking yields robust query interpretation, including cross-dataset transfer to Pascal VOC without retraining.

\section{Related Work}
\label{sec:related_work}

\paragraph{Interactive and Conversational XAI:}This challenge highlights the "social" nature of explanations~\cite{miller2019explanation}, suggesting they should be interactive dialogues rather than static artifacts. While early XAI systems categorized user intent into queries like Why? or What if?~\cite{liao2020questioning}, and visual tools like Gamut~\cite{hohman2019gamut} or the What-If Tool~\cite{wexler2019if} enabled manual counterfactual inspection, these interfaces often require significant domain expertise. Conversational XAI lowers this barrier by allowing natural language interaction; while systems like TalkToModel~\cite{slack2022talktomodel} pioneered this for tabular data, our work extends this paradigm to global vision explanations, supporting complex structural queries over necessary and sufficient conditions.

\paragraph{LLMs as Neuro-Symbolic Interpreters:}To bridge the gap between natural language and formal logic, we leverage Large Language Models (LLMs) as neuro-symbolic interpreters rather than direct explanation generators, which are prone to hallucination. Existing work uses LLMs as semantic parsers (e.g., text-to-SQL~\cite{yu2018spider}~\cite{scholak2021picard}) or tool-augmented routers~\cite{schick2023toolformer}. Within XAI, LLMs have been used to describe neurons~\cite{bills2023language} or retrieve static artifacts~\cite{singh2024rethinking}. In contrast, GLARE treats the LLM as a logic-constrained parser that maps user intent to a deterministic "grammar of explanations" via verifiable SQL templates. This ensures that the flexibility of natural language is grounded in formal correctness, enabling precise logical aggregations and counterfactual queries over the model’s reasoning structure.
\section{Methodology}
\label{sec:methodology}

We present a natural language interface for querying global explanations
of image classifiers. Our system builds upon the global explanation
framework of Vasu et al.~\cite{vasu2026local}, which generates
concept-based explanations for black-box image classifiers, expressed
as Disjunctive Normal Form (DNF) formulas that describe important
object combinations for each class. Our system enables users to pose
analytical questions in natural language and receive structured,
interpretable answers along with supporting evidence images. More formally, let $f_\theta: \mathcal{X} \rightarrow \mathcal{Y}$ denote an image classifier mapping images $x \in \mathcal{X}$ to labels $y \in \mathcal{Y}$. We assume a dataset $\mathcal{D} = \{(x_i, y_i)\}_{i=1}^N$ and predictions $\hat{y}_i = f_\theta(x_i)$. We assume access to a local explanation generator $E$ producing an explanation artifact $e_i = E(x_i, f_\theta)$ for each input, where $E$ is an the result of a concept based attribution methods or prototypes. Users pose natural-language questions $q$ about aggregate model behavior over subsets of $\mathcal{D}$ (e.g., by class, confusion pair, or attribute). We aim to answer $q$ by retrieving and aggregating relevant local explanation artifacts $\{e_i\}$.

\subsection{System Overview}
\label{sec:overview}

\begin{figure*}[t]
    \centering
    \includegraphics[width=\textwidth]{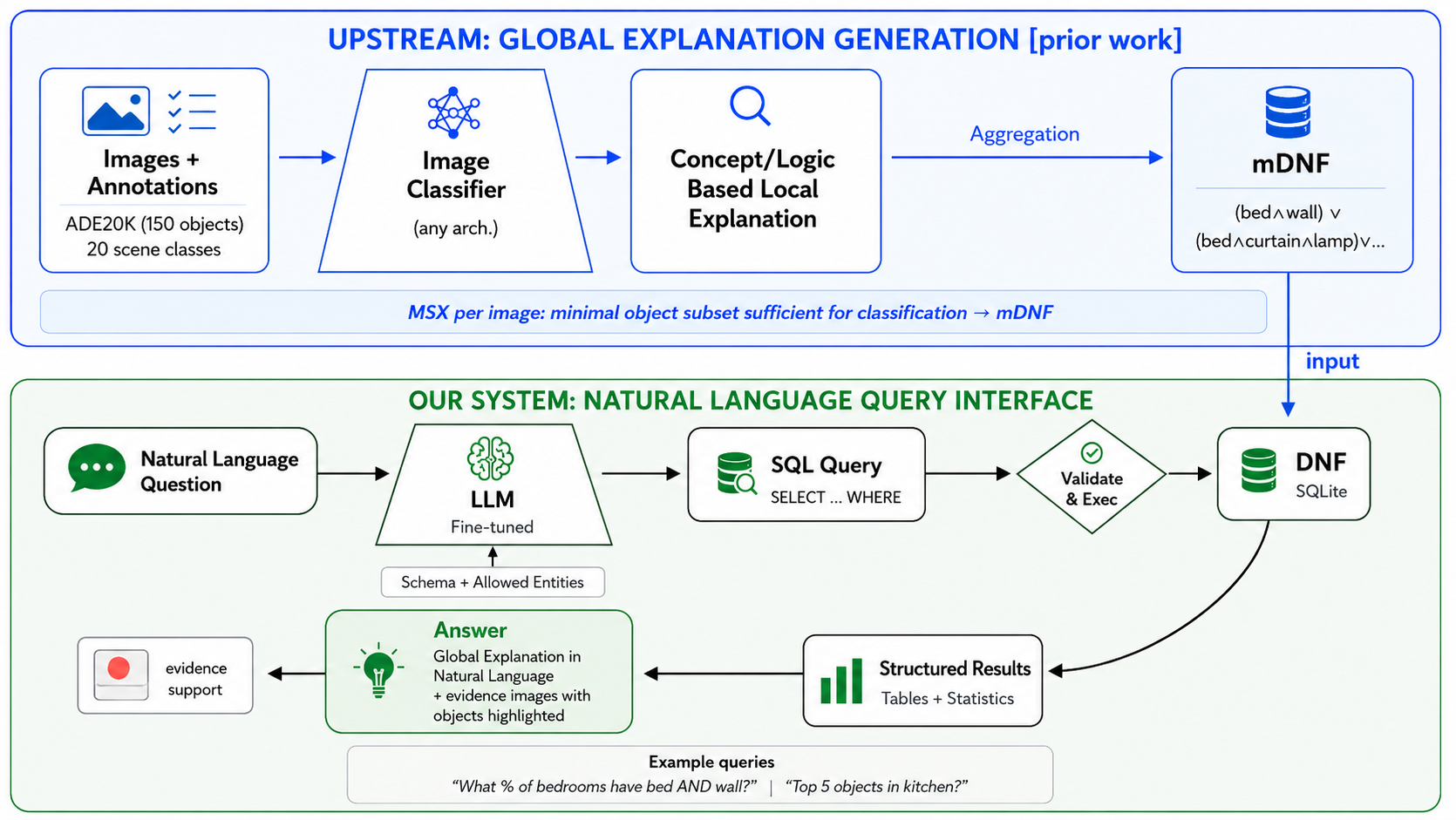}
    \caption{
    End-to-end pipeline. Top: The upstream framework  generates mDNF explanations from aggregation of local concept or logic based explanation. Bottom: Our system translates natural language questions into validated SQL queries executed against the explanation database, returning structured answers with supporting evidence images.
    }
    \label{fig:full_pipeline}
\end{figure*}

Our system follows a \emph{parse-validate-execute} pipeline (Figure~\ref{fig:full_pipeline}). A user poses a question in natural language (e.g., \emph{``What percentage of bedroom images contain both bed and wall?''}). Our LLM-based query interpreter translates the question into a structured SQL query by selecting from a set of predefined query templates and extracting the relevant parameters (class names, object names, thresholds). The parameterized template is instantiated as an executable SQL query, validated for correctness and safety, then executed against a database encoding the global explanations. Results are returned as structured data along with supporting evidence images highlighting the relevant objects. The structured data returned is then converted back to natural language using a generic small LLM of the same size. Images illustrating the supporting local explanations, which highlight contributing regions based on the original image with a segmentation map as shown in Figure \ref{fig:evidence}.

Because the model learns to generate SQL over a fixed relational schema, it acquires the compositional structure of the query language itself and not merely a mapping from phrases to template indices. This enables generalization along multiple axes: to unseen entity combinations, linguistic variations, novel compositions of known SQL fragments, and even entirely new datasets sharing the same schema. At the same time, anchoring generation in predefined templates mitigates risks of malformed or semantically incorrect queries while preserving the expressiveness needed for meaningful explanation interrogation. Given a query, the interface returns:(1) natural-language summary with relevant statistics; (2) supporting local explanations (examples); (3) visualizations aligned with query intent.

\subsection{LLM-Based Query Interpretation}
\label{sec:interpretation}

We formulate natural-language query interpretation as a structured SQL generation problem: given a user query $q$, the model selects an appropriate query pattern and extracts the parameters $\phi(q)$ needed to instantiate it as executable SQL. We fine-tune Gemma 2-9B~\cite{riviere2024gemma} using Low-Rank Adaptation (LoRA)~\cite{Hu2021LoRALA} with 4-bit quantization (QLoRA) with a LoRA rank: 16, alpha: 32, dropout: 0.05.

\paragraph{Training Data Generation.}
We generate synthetic training data covering 24 distinct query templates (Section~\ref{sec:templates}), producing 50,000 training and 2,000 validation examples. For each template, we sample random class-object combinations and apply natural language variation: (1) \textbf{Synonym substitution}: Operators (\emph{``and''}, \emph{``\&''}, \emph{``together with''}), quantifiers (\emph{``percentage''}, \emph{``\%''}, \emph{``proportion''}), ranking terms (\emph{``top''}, \emph{``most common''}, \emph{``leading''}). (2)\textbf{Phrasing templates}: Natural language templates per query type (e.g., \emph{``What \% of X have Y?''} vs. \emph{``What proportion of X contain Y?''}). During fine-tuning, we employ a custom collator (\texttt{SqlFenceCollator}) that masks the training loss to \emph{only} tokens between \texttt{SQL\_START} and \texttt{SQL\_END}. 

\subsection{Query Templates}
\label{sec:templates}

We define 24 query templates corresponding to common analytical tasks over global explanations. Each template captures a specific question type and is parameterized by entities extracted from the user query (target class, object names, comparison class, thresholds, etc.). The templates are organized into three tiers of increasing complexity: \emph{Core} queries cover fundamental object-class relationships such as frequency, boolean combinations, top-$k$ ranking, co-occurrence, and class ranking; \emph{Extended} queries leverage SQL features like self-joins for $N$-way combinations, cross-class comparisons, set operations, conditional co-occurrence, and confidence-filtered analysis; and \emph{Contrastive} queries enable counterfactual analysis including absence analysis, threshold queries, and distinguishing features between classes. The full taxonomy with all 24 templates and the prompt structure is provided in the supplementary material. Crucially, the template set is not a fixed system boundary: adding a new question type requires only defining a new SQL pattern and regenerating synthetic training data, after which the entire pipeline, data generation, fine-tuning, and evaluation runs automatically without manual annotation.

\section{Experimental Setup}
\label{sec:experiments}

We evaluate GLARE along four axes: (i)~in-distribution accuracy on held-out queries over the training dataset, (ii)~robustness to natural-language perturbations, (iii)~out-of-distribution generalization to novel phrasing and unseen SQL constructs, and (iv)~cross-dataset transfer to an entirely different object vocabulary and scene taxonomy.

\subsection{Datasets}
\label{sec:datasets}

\paragraph{Training data.}
Training examples are generated synthetically by sampling from the 24 query templates (Section~\ref{sec:templates}).
Each template produces a (natural-language question, SQL query) pair by randomly selecting objects and scene classes from the ADE20K vocabulary of 150~objects and 35~scene categories.
We generate 50{,}000 training pairs (seed\,=\,42) and 2{,}000 validation pairs (seed\,=\,1{,}042), each formatted as chat-style messages with the system prompt, user question, and gold SQL delimited by \texttt{SQL\_START}/\texttt{SQL\_END} markers.
The training set uses only object and class \emph{name lists}; no data from the ground-truth explanation database is accessed during training. The reference database is built from the mDNF explanations~\cite{vasu2026local} of a VGG19 model computed over ADE20K~\cite{zhou2017scene} dataset ~\cite{vasu2026local}. The local explanations achieves a Fidelity$^+$ and Fidelity$^-$ of $0.113\pm0.297$ and $0.992\pm0.090$ respectively, reflecting their faithfulness to the underlying model. Note, our system can accommodate any local explanation algorithm that generates explanations in symbolic form. It serves exclusively as the \emph{execution environment} during evaluation: generated SQL queries are executed against this database to verify the correctness.

\paragraph{Evaluation splits:}We construct three evaluation sets that show progressively more challenging forms of generalization:

\begin{enumerate}

  \item \textbf{OOD Test} (300 examples):
        Split into \emph{phrasing variations} of trained templates (182~examples) and \emph{novel SQL constructs} absent from training (118~examples).
        Phrasing variations include informal language (e.g., ``yo how many''), negations, double negations, ambiguous wording, alias usage, extreme boundary values, and multi-step reasoning.
        Novel SQL constructs include window functions (\texttt{RANK() OVER}), \texttt{CASE} expressions, correlated subqueries, complex \texttt{HAVING} clauses, XOR logic, relative comparisons, and cross-class set differences, none of which appear in any training template. This split directly tests whether the model has learned generalizable SQL structure beyond the specific patterns it was trained on.

  \item \textbf{Robustness Test} (500 examples):
        Each Fresh Test query is subjected to seven perturbation types: \emph{spelling} errors, \emph{grammar} corruptions, \emph{synonym} substitution, \emph{verbose} padding, \emph{telegraphic} compression, \emph{word drop}, and \emph{word swap}.
        We additionally measure \emph{consistency} (whether paraphrases of the same query yield identical results) and \emph{sanity} (whether outputs satisfy domain constraints).

  \item \textbf{Pascal~VOC} (500 examples):
        A cross-dataset evaluation in which the model, trained exclusively on ADE20K entity names, is tested against a database built from Pascal~VOC~\cite{chen2014detect} annotations with a disjoint object vocabulary (166~objects, different scene taxonomy).
        Test queries follow the same template distribution. This evaluates whether the learned SQL structure transfers to an entirely new domain without retraining.
\end{enumerate}

\subsection{Models}
\label{sec:models}

To assess the impact of fine-tuning and the generality of our approach across model scales, we evaluate two instruction-tuned model families:

\begin{itemize}
  \item \textbf{Gemma~2}~\cite{riviere2024gemma}: 2B\,/\,9B\,/\,27B parameters
  \item \textbf{Qwen~2.5}~\cite{qwen_qwen25_2025}: 0.5B\,/\,7B\,/\,14B parameters
\end{itemize}

\noindent All fine-tuned models use the identical QLoRA configuration and SQL-fence loss masking described in Section~\ref{sec:interpretation}, with the same 50{,}000 synthetic training examples. We additionally evaluate each model in its base (untrained) configuration,
while keeping all other aspects the same to isolate the contribution of our synthetic training pipeline (Section~\ref{sec:results_fresh}).

\subsection{Evaluation Metrics}
\label{sec:eval_metrics}

We report the following metrics, evaluated by executing generated SQL against the ground-truth database and comparing the returned result sets:

\begin{itemize}
  \item \textbf{Fence Detection (\%)}: Fraction of outputs correctly delimited by \texttt{SQL\_START}/\texttt{SQL\_END} markers, confirming adherence to the expected output format.

  \item \textbf{SQL Parse Rate (\%)}: Fraction of outputs containing syntactically valid SQL.

  \item \textbf{Execution Rate (\%)}: Fraction of SQL queries that execute without runtime error against the explanation database.

  \item \textbf{Result Match (\%)}: Fraction of queries whose executed results match the gold-standard output.
  We use \emph{relaxed} matching: row-order and \texttt{LIMIT} differences are tolerated, extra columns are permitted, and numeric values are compared with 1\% relative tolerance. This reflects the principle that semantically equivalent SQL may differ in surface form.

  \item \textbf{Partial Match (\%)}: For queries that fail exact match, we compute the Jaccard similarity over first-column value sets. Queries with Jaccard $> 0.5$ are counted as partial matches, indicating substantially correct query intent despite structural differences.
\end{itemize}

\noindent The robustness evaluation (Section~\ref{sec:results_robustness}) additionally reports:

\begin{itemize}

\item \textbf{Robustness Score (\%)}: The percentage of the unperturbed baseline result-match rate preserved under each perturbation type. A score of $100\%$ indicates zero degradation.

  \item \textbf{Consistency Rate (\%)}: For each baseline query that executes successfully, two paraphrases are generated (via synonym substitution and verbose rephrasing). Consistency is the fraction of paraphrases that produce result sets identical to the original.

\end{itemize}
\section{Results and Discussion}
\label{sec:results}

\begin{table}[t]
\centering
\caption{Performance on the ADE20K Fresh Test set (500 examples). Fine-tuned (FT) models use QLoRA with SQL-fence loss masking on 50{,}000 synthetic training pairs generated from 24 query templates. Base models receive the same prompt but no task-specific fine-tuning. Cross rows denote zero-shot transfer performance to a different taxonomy from PASCAL VOC (166 objects) while 'Regex' uses Regular Expression (Regex) for a non-learning baseline.}
\label{tab:merged_results}
\small
\setlength{\tabcolsep}{4pt}
\begin{tabular}{ll c ccc ccc}
\toprule
\multirow{2}{*}{\textbf{Metric}} & \multirow{2}{*}{\textbf{Setting}} & \textbf{Regex} & \multicolumn{3}{c}{\textbf{Gemma 2}} & \multicolumn{3}{c}{\textbf{Qwen 2.5}} \\
\cmidrule(lr){4-6} \cmidrule(lr){7-9}
& & \textbf{Baseline} & \textbf{2B} & \textbf{9B} & \textbf{27B} & \textbf{0.5B} & \textbf{7B} & \textbf{14B} \\
\midrule
\multirow{3}{*}{\textbf{Fence}} 
& base  & \multirow{3}{*}{100.0} & 71.2  & 100.0 & 100.0 & 41.4 & 81.6  & 66.2 \\
& Fine-tuned    &                        & 100.0 & 100.0 & 100.0 & 93.8 & 100.0 & 100.0 \\
& Cross &                        & 100.0 & 100.0 & 100.0 & 93.8 & 100.0 & 100.0 \\
\midrule
\multirow{3}{*}{\textbf{Parse}} 
& base  & \multirow{3}{*}{100.0} & 6.2   & 3.2   & 0.6   & 2.2  & 37.8  & 15.6 \\
&  Fine-tuned    &                        & 100.0 & 100.0 & 100.0 & 4.6  & 97.6  & 100.0 \\
& Cross &                        & 100.0 & 100.0 & 100.0 & 4.6  & 98.4  & 99.4 \\
\midrule
\multirow{3}{*}{\textbf{Exec.}} 
& base  & \multirow{3}{*}{100.0} & 6.2   & 3.2   & 0.6   & 2.2  & 37.8  & 15.6 \\
&  Fine-tuned    &                        & 100.0 & 100.0 & 100.0 & 4.6  & 97.6  & 100.0 \\
& Cross &                        & 100.0 & 100.0 & 100.0 & 4.6  & 98.4  & 99.4 \\
\midrule
\multirow{3}{*}{\textbf{Match}} 
& base  & \multirow{3}{*}{74.4}  & 0.0   & 0.2   & 0.0   & 0.0  & 3.2   & 4.4 \\
&  Fine-tuned    &                        & 95.4  & 95.2  & 95.2  & 4.4  & 93.0  & 95.4 \\
& Cross &                        & 90.0  & 89.6  & 90.6  & 4.4  & 87.2  & 90.0 \\
\midrule
\multirow{3}{*}{\textbf{Partial}} 
& base  & \multirow{3}{*}{5.6}   & 0.4   & 0.4   & 0.2   & 0.4  & 3.4   & 2.0 \\
&  Fine-tuned    &                        & 2.8   & 2.4   & 3.0   & 0.0  & 3.0   & 2.6 \\
& Cross &                        & 1.0   & 3.4   & 1.0   & 0.0  & 3.2   & 2.8 \\
\bottomrule
\end{tabular}
\end{table}
\subsection{In-Distribution Query Accuracy}
\label{sec:results_fresh}

Table~\ref{tab:merged_results} summarizes performance on the 500-example Fresh Test set, comparing both fine-tuned and base (untrained) model configurations.

Among fine-tuned models, Gemma~2 9B achieves perfect scores on all three structural metrics (fence detection, SQL parse, and execution) and a result-match rate of 95.2\%, with an additional 2.4\% of examples achieving partial matches (Jaccard $> 0.5$), bringing the effective accuracy above 97\%.
Gemma~2 27B and Gemma~2 2B achieve comparable result-match rates of 95.2\% and 95.4\% respectively, both with perfect structural metrics, indicating that the fine-tuning pipeline saturates in-distribution performance across a wide range of model scales within the Gemma~2 family. While the non-learning Regex baseline achieves perfect structural metrics, our fine-tuned models substantially outperform its 74.4\% result-match rate, achieving over 95\% match accuracy across the Gemma 2 family.
Base models, despite receiving the identical prompt, achieve near-zero result-match accuracy (Table~\ref{tab:merged_results}), confirming that the observed generalization is the product of task-specific fine-tuning rather than pre-existing SQL knowledge or in-context learning.
Across architectures, a minimum capacity threshold is required: Qwen~2.5 0.5B fails even after fine-tuning, while Qwen~2.5 7B reaches 93.0\%. Please refer to the Appendix \ref{sec:results_fresh} for the breakdown of performance across different query types.
 
\subsection{Robustness to Input Perturbations}
\label{sec:results_robustness}

Table~\ref{tab:robustness} reports robustness to seven perturbation types applied to the 500-example Fresh Test set, along with consistency and sanity metrics, explicitly comparing the fine-tuned Gemma~2 9B model against the regex baseline.

The fine-tuned model demonstrates significant architectural advantages over regex, particularly on perturbations that reflect authentic user behavior:

\textbf{Spelling errors:} The model maintains a 79.6\% match rate compared to the regex's 48.7\%, a stark +31 percentage point (pp) gap. Real users frequently mistype words (e.g., ``bedrrom'', ``kitchn'', ``chiar''). A regex requires the canonical token verbatim and offers no path forward without engineering a separate fuzzy-matching system, whereas the model natively absorbs typographical noise. This +31~pp gap is arguably the cleanest evidence in the suite demonstrating the superiority of fine-tuning. \textbf{Synonym substitution:} The model exhibits zero degradation (100\% robustness) to synonyms, outperforming the regex by +17~pp in absolute match rate. Users naturally express concepts using diverse vocabulary (e.g., ``proportion'', ``share'', ``fraction'', ``slice of the pie''). The regex only recognizes hand-written synonyms; the moment a user employs an unmapped term, it silently falls through to an incorrect template. The model, conversely, leverages broad lexical knowledge from pretraining to infer meaning and degrade gracefully. \textbf{Word drop:} Real users often submit terse, ungrammatical queries (e.g., ``percent bedroom bed'', ``top chair classroom''). While word drop is the most damaging perturbation overall, removing core content words inherently destroys the information needed to generate correct SQL, the model still maintains a +2.7~pp advantage (42.6\% vs. 39.9\%). The regex strictly requires trigger tokens to be present and adjacent to fire its rules. The model's ability to infer intent from partial context is the only scalable path forward, as simply writing more regex rules cannot recover missing tokens.

The other four perturbations (verbose padding, telegraphic compression, grammar variations, and synonym substitutions) are primarily designed to test paraphrase invariance, varying surface structure while preserving core content words. While the regex baseline scores artificially high on some of these metrics, a regex literally ignores surface structure, so by construction, it cannot be confused by it. This is not robustness in the language-understanding sense; it is merely deafness to syntax. Because of this architectural blindness, direct comparisons to the regex baseline on syntactic paraphrase metrics are fundamentally uninformative.

Finally, model \textbf{consistency} across paraphrases reaches 94.1\% (744/791~tests), indicating that semantically equivalent questions reliably yield identical SQL, a critical property for user trust. Furthermore, all 1{,}485 \textbf{sanity checks} pass (100\%), confirming that generated results never violate underlying domain constraints (such as percentages outside $[0,100]$ or negative counts).

\begin{table}[t]
\centering
\caption{Robustness evaluation (Gemma~2 9B and Regex Baseline, Fresh Test, 500 examples). Robustness score is the percentage of the unperturbed baseline match rate preserved under each perturbation.}
\label{tab:robustness}
\small
\setlength{\tabcolsep}{3pt}
\begin{tabular}{lr cc cc cc cc}
\toprule
\multirow{2}{*}{\textbf{Perturbation}} & \multirow{2}{*}{$n$} & \multicolumn{2}{c}{\textbf{Prs.}} & \multicolumn{2}{c}{\textbf{Exec.}} & \multicolumn{2}{c}{\textbf{Match}} & \multicolumn{2}{c}{\textbf{Rob.\,(\%)}} \\
\cmidrule(lr){3-4} \cmidrule(lr){5-6} \cmidrule(lr){7-8} \cmidrule(lr){9-10}
& & \textbf{Orig.} & \textbf{Regex} & \textbf{Orig.} & \textbf{Regex} & \textbf{Orig.} & \textbf{Regex} & \textbf{Orig.} & \textbf{Regex} \\
\midrule
Synonym      & 150 & 97.3 & 100.0 & 97.3 & 100.0 & 89.3 & 72.0 & 100 & 89 \\
Verbose      & 500 & 98.4 & 100.0 & 98.4 & 100.0 & 86.6 & 80.8 & 97 & 100 \\
Spelling     & 275 & 98.9 & 100.0 & 98.9 & 100.0 & 79.6 & 48.7 & 89 & 60 \\
Telegraphic  & 450 & 99.1 & 100.0 & 99.1 & 100.0 & 75.3 & 74.9 & 84 & 93 \\
Word swap    & 384 & 98.7 & 100.0 & 98.7 & 100.0 & 73.4 & 67.2 & 82 & 83 \\
Grammar      & 140 & 98.6 & 100.0 & 98.6 & 100.0 & 67.9 & 84.3 & 76 & 104 \\
Word drop    & 303 & 96.7 & 100.0 & 96.7 & 100.0 & 42.6 & 39.9 & 48 & 49 \\
\midrule
\multicolumn{8}{l}{Consistency (paraphrase $\rightarrow$ same result)} & 94.1 & 95.0 \\
\multicolumn{8}{l}{Sanity (domain constraints satisfied)} & 100.0 & 100.0 \\
\bottomrule
\end{tabular}
\end{table}
\subsection{Out-of-Distribution Generalization}
\label{sec:results_ood}

The OOD evaluation (Table~\ref{tab:ood_results}) probes two distinct aspects of generalization: resilience to novel natural-language phrasing of trained query patterns, and the ability to produce SQL constructs never seen during training.
Across all 300 examples, the model maintains 99.3\% parse and execution rates, indicating that the learned SQL structure generalizes broadly even when semantics diverge from training.

\paragraph{Phrasing variations.}
On the 182~examples that rephrase trained query types using unfamiliar surface forms, Gemma~2 9B achieves 45.1\% exact match.
Performance varies widely: \texttt{zero\_threshold} (100\%) and \texttt{nested\_question} (90\%) are handled well, the model correctly interprets complex nested clause structure and edge-case thresholds.
\texttt{negation} (70\%) shows reasonable handling of NOT-style queries.
However, \texttt{informal\_question} (12\%), \texttt{double\_negation} (0\%), and \texttt{comparative} (0\%) reveal brittleness: the model has learned the underlying SQL patterns but struggles when the surface form deviates substantially from training templates.

\paragraph{Novel SQL constructs.}
On the 118~examples requiring SQL constructs entirely absent from training, the model achieves 19.5\% exact match.
\texttt{chained\_filter} (100\%) and \texttt{string\_pattern} (100\%) represent compositional generalization i.e the model assembles familiar SQL fragments (\texttt{WHERE} clauses, \texttt{LIKE} operators) into structures it was never explicitly trained on.
Conversely, \texttt{window\_function}, \texttt{case\_expression}, \texttt{subquery\_select}, \texttt{having\_complex}, and \texttt{relative\_comparison} all score 0\%, confirming that the model learns the compositional structure of the SQL it is trained on, but cannot extrapolate to truly novel syntax (e.g., \texttt{RANK() OVER}, \texttt{CASE WHEN}).
Notably, even on these unsupported constructs, the model maintains 100\% execution rate by producing the closest known template rather than generating invalid SQL, a form of graceful degradation that means expanding the system's analytical scope requires only adding new templates to the synthetic training pipeline.

\begin{table}[t]
\centering
\caption{Out-of-distribution results (Gemma~2 9B, 300 examples). Left: phrasing variations of trained query patterns. Right: novel SQL constructs absent from training.}
\label{tab:ood_results}
\footnotesize
\setlength{\tabcolsep}{2pt}
\begin{tabular}{@{}lrcc@{\hskip 8pt}lrcc@{}}
\toprule
\multicolumn{4}{@{}c@{\hskip 8pt}}{\textit{Phrasing Variations (trained)}} & \multicolumn{4}{c@{}}{\textit{Novel SQL Constructs (untrained)}} \\[2pt]
\textbf{Query Type} & $n$ & \textbf{Ex} & \textbf{Ma} & \textbf{Query Type} & $n$ & \textbf{Ex} & \textbf{Ma} \\
\midrule
zero\_threshold    & 11 & 100 & 100 & chained\_filter      & 12 & 100 & 100 \\
nested\_question   & 29 &  97 &  90 & string\_pattern      &  8 & 100 & 100 \\
negation           & 23 & 100 &  70 & never\_appears       & 10 & 100 &  20 \\
extreme\_topk      & 25 & 100 &  52 & xor                  &  9 & 100 &  11 \\
ambiguous          & 13 & 100 &  38 & case\_expression     & 20 & 100 &   0 \\
alias              & 17 & 100 &  35 & subquery\_select     & 17 & 100 &   0 \\
informal\_question & 34 & 100 &  12 & window\_function     & 15 & 100 &   0 \\
multistep          & 13 & 100 &   8 & relative\_comparison & 14 &  93 &   0 \\
comparative        & 10 & 100 &   0 & having\_complex      & 11 & 100 &   0 \\
double\_negation   &  7 & 100 &   0 & cross\_class\_object &  2 & 100 &   0 \\
\midrule
\textbf{Subtotal} (182) & & \textbf{99.5} & \textbf{45.1} & \textbf{Subtotal} (118) & & \textbf{99.2} & \textbf{19.5} \\
\bottomrule
\multicolumn{8}{c}{\textbf{Overall} (300):\enspace Exec \textbf{99.3},\enspace Match \textbf{35.0}} \\
\end{tabular}
\end{table}

\subsection{Cross-Dataset Transfer}
\label{sec:results_pascal}
Table~\ref{tab:merged_results} (row Cross) presents a cross-dataset evaluation where the model, trained exclusively on ADE20K entity names, is tested on a database built from Pascal~VOC annotations, a different object vocabulary (166~objects vs.\ 150) and scene taxonomy.
Gemma~2 9B achieves 89.6\% result-match accuracy with perfect structural metrics (100\% fence detection, parse, and execution).
Gemma~2 27B achieves the highest result-match accuracy at 90.6\%, while Gemma~2 2B reaches 90.0\%, both slightly outperforming the 9B variant.
Performance is tightly clustered across all three Gemma~2 scales, suggesting that the learned SQL structure transfers robustly even at 2B parameters.
Performance is strong across all query types, with 17 of 25 types at ${\geq}$95\% accuracy.
These results provide the strongest evidence that the model has learned generalizable \emph{relational structure} and SQL's compositionality separates query structure from vocabulary, enabling deployment on any new explanation database conforming to the same schema without retraining, given only an entity name list.

\subsection{Cross-Model Comparison}
\label{sec:results_cross_model}

Table~\ref{tab:model_comparison} summarizes fine-tuned model performance across all evaluation axes.
Gemma~2 9B and Qwen~2.5 7B achieve comparable Fresh Test accuracy (95.2\% vs.\ 93.0\%) despite being architecturally distinct model families, Gemma~2 and Qwen~2.5 differ in pre-training data, tokenizer, and architectural details such as attention head configuration.
Gemma~2 27B and Gemma~2 2B achieve 95.2\% and 95.4\% respectively, confirming that in-distribution performance saturates across model scales within the Gemma~2 family.
This close agreement confirms that our synthetic training pipeline and SQL-fence loss masking are \emph{architecture-agnostic}: any sufficiently large instruction-tuned causal language model can serve as the backbone.

On OOD generalization, Gemma~2 27B achieves 34.3\%, comparable to Gemma~2 9B (35.0\%), while Gemma~2 2B achieves 30.3\%, suggesting that OOD performance does not scale as strongly with model size as in-distribution accuracy.
Qwen~2.5 7B slightly outperforms all Gemma~2 variants on OOD (40.0\%), suggesting that OOD performance may depend on the base model's pre-training mixture as much as on fine-tuning.
On cross-dataset transfer, Gemma~2 27B achieves 90.6\% on Pascal~VOC, with Gemma~2 2B at 90.0\%, both slightly outperforming the 9B variant (89.6\%).
At the lower end, Qwen~2.5 0.5B fails entirely (4.4\% Fresh, 4.7\% OOD), confirming that a minimum model capacity is required for the fine-tuning pipeline to take effect.

\begin{table}[t]
\centering
\caption{Cross-model comparison of fine-tuned models across all evaluation axes. Fresh and Pascal report result-match (\%); OOD reports overall match (\%); Robust reports the mean robustness weighted average score (\%) across all seven perturbation types.}

\label{tab:model_comparison}
\small
\begin{tabular}{llcccc}
\toprule
\textbf{Model} & \textbf{Params} & \textbf{Fresh} & \textbf{OOD} & \textbf{Pascal} & \textbf{Robust} \\
\midrule
Gemma~2     & 2B  & 95.4 & 30.3 & 90 & 78.9 \\
Gemma~2     & 9B  & 95.2 & 35.0 & 89.6 & 82.8 \\
Gemma~2     & 27B & 95.2 & 34.3 & 90.6 & 85.9 \\
\midrule
Qwen~2.5    & 0.5B & 4.4 & 4.7 & 4.4 & 54.4 \\
Qwen~2.5    & 7B  & 93.0 & 40.0 & 87.2 & 82.2 \\
Qwen~2.5    & 14B & 95.4 & 40.0 & 90.0 & 82.5 \\
\bottomrule
\end{tabular}
\end{table}

\subsection{Qualitative Case Study}
\label{sec:case_studies}

We present representative examples drawn from actual evaluation
outputs, illustrating system behavior across the generalization
spectrum.  For each case we show the user's natural-language question,
the model-generated SQL, the executed result,
and the natural-language answer returned to the user by
the downstream NL generation stage.
Where noted, SQL is abbreviated to highlight the structurally
relevant clauses.

\paragraph{Case Study: End-to-end multi-modal response co-occurrence
with visual grounding.}
\begin{quote}
\textbf{Question:} ``In living\_room, what objects appear with
sculpture?''
\end{quote}

\noindent\textbf{NL Output:} \emph{``Sculptures in living rooms most
often appear with walls (80\% of the time), followed by sofas (60\%).
Other common objects found alongside sculptures include cushions,
coffee tables, and windowpanes.''}

\noindent\textbf{Top co-occurrences} (count / \% of anchor images):
\texttt{wall}~12/80\%,
\texttt{sofa}~9/60\%,
\texttt{cushion}~8/53\%,
\texttt{coffee\_table}~7/47\%,
\texttt{windowpane}~5/33\%,
\texttt{floor}~4/27\%,
\texttt{painting}~4/27\%,
\texttt{ceiling}~3/20\%,
\texttt{curtain}~3/20\%,
\texttt{fireplace}~3/20\%.

\begin{figure}[h!]
    \centering
    \includegraphics[width=0.9\linewidth]{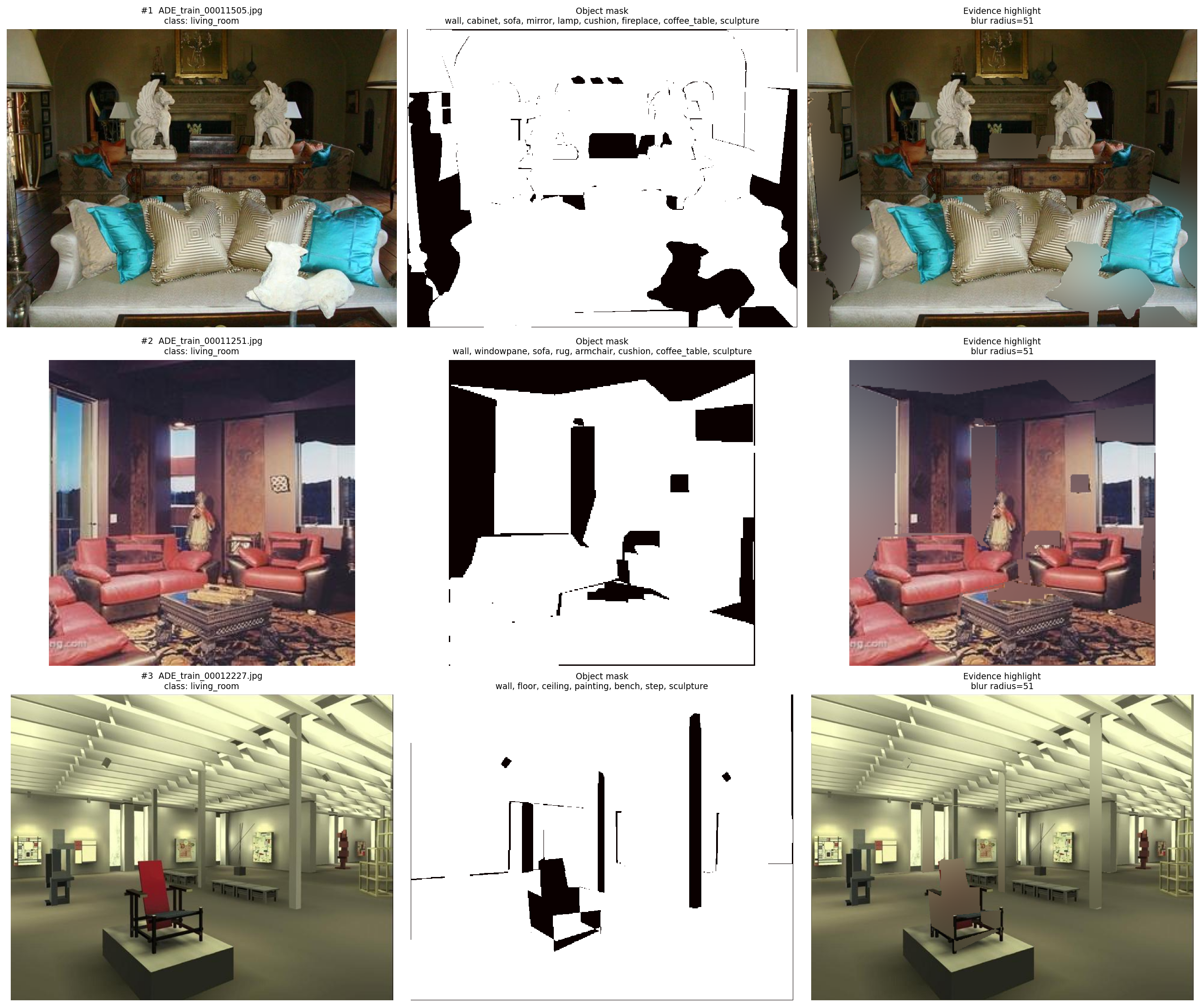}
    \caption{Visual grounding evidence for \textit{``In living rooms, what
    objects appear with sculpture?''}: top-3 evidence images showing the original image (Column 1), Objects deemed important (with value 1) by local explanations (Column 2), and finally the masked image highlighting only the important objects.}
    \label{fig:evidence}
\end{figure}

The system also provides visual grounding
(Figure~\ref{fig:evidence}): the top-3 evidence images with
relevant objects identified, enabling users to verify that
co-occurrence statistics correspond to meaningful patterns.
This case demonstrates the full explanatory loop: a natural-language
question translated into SQL, executed, and returned as a
multi-modal response comprising a fluent summary, quantitative
statistics, and visual evidence.

\section{Conclusion and Future Work}
\label{sec:conclusion_limit}

We have presented GLARE, an LLM-mediated interface that transforms global explanation consumption from the passive inspection of static artifacts into an active, query-driven dialogue. By constraining generation to a SQL intermediate representation over a fixed relational schema, the system ensures compositional expressiveness and formal correctness while remaining accessible to users without programming expertise.
Our synthetic training pipeline and fence-masked fine-tuning are architecture-agnostic, requiring no manual annotation and enabling zero-shot transfer to new datasets provided only with an entity name list.
Empirically, small fine-tuned models ($\geq$7\,B parameters) achieve ${>}$95\% accuracy on in-distribution queries, transfer to unseen domains with ${\sim}$90\% accuracy, and maintain high robustness to linguistic perturbations.

One of the future works includes dealing with the template taxonomy that is extensible but not yet exhaustive; covering additional SQL constructs requires only additions to the synthetic generator.
Overall, these results suggest that LLM-mediated querying offers a practical and reliable path toward more accessible, human-centered global explanations in XAI workflows.

\bibliographystyle{splncs04}
\bibliography{ref}
\newpage
\appendix
\section{Appendix}
\subsection{Query type breakdown}

Table~\ref{tab:fresh_breakdown} provides a per-query-type breakdown for Gemma~2 9B. The model achieves 100\% accuracy on 18 of 25 evaluated query types, spanning percentage calculations, top-$k$ ranking, co-occurrence joins, set operations, conditional existence checks, threshold filtering, and statistical aggregations. In contrast, the non-learning Regex baseline proves highly brittle on queries requiring complex structural constraints, suffering severe accuracy drops on types like \texttt{combos} (8\%), \texttt{absence\_count} (36\%), and \texttt{topclass} (40\%). The two underperforming categories for the model, \texttt{combos} (49\%) and \texttt{images\_with\_exact\_count} (43\%), involve complex multi-way self-joins and exact-count \texttt{HAVING} clauses, respectively.
Notably, the model achieves ${\geq}$98\% on 56~examples labeled \texttt{unknown} (edge-case queries outside the standard template taxonomy), suggesting robustness to minor distribution shifts even within the in-distribution evaluation.
These results confirm that the model internalizes the compositional structure of SQL, joins, filters, aggregations, subqueries as reusable rules that can be instantiated with arbitrary entity names, rather than memorizing specific query-answer associations. 

\begin{table}[h!]
\centering
\caption{Per-query-type result-match accuracy (\%) on the Fresh Test set for Gemma~2 9B and the Regex (RX) baseline, sorted by example count.}
\label{tab:fresh_breakdown}
\footnotesize
\setlength{\tabcolsep}{4pt} 
\begin{minipage}[t]{0.48\columnwidth} 
\centering
\begin{tabular}{lrcc}
\toprule
\textbf{Query Type} & $n$ & \textbf{Gemma} & \textbf{RX} \\
\midrule
percent\_simple          & 70 & 99  & 67 \\
cooccur                  & 65 & 100 & 92 \\
topk\_objects            & 37 & 100 & 81 \\
percent\_exclude         & 36 & 100 & 90 \\
combos                   & 35 & 49  & 8  \\
topclass                 & 29 & 100 & 40 \\
percent\_prob.           & 27 & 100 & 83 \\
cross\_cls\_comp.        & 15 & 100 & 100 \\
set\_difference          & 13 & 100 & 100 \\
obj\_per\_img\_st.       & 13 & 100 & 100 \\
threshold\_qry           & 11 & 100 & 100 \\
cnt\_distinct\_obj       & 11 & 100 & 100 \\
\bottomrule
\end{tabular}
\end{minipage}%
\hfill
\begin{minipage}[t]{0.48\columnwidth}
\centering
\begin{tabular}{lrcc}
\toprule
\textbf{Query Type} & $n$ & \textbf{Gemma} & \textbf{RX} \\
\midrule
percent\_and             & 10 & 100 & 100 \\
existence\_chk           &  8 & 100 & 80 \\
least\_common            &  8 & 100 & 89 \\
cond\_cooccur            &  7 & 100 & 100 \\
absence\_count           &  7 & 100 & 36 \\
class\_count             &  7 & 100 & 54 \\
set\_intersect.          &  7 & 100 & 100 \\
img\_w\_exact\_cnt       &  7 & 43  & 100 \\
percent\_or              &  7 & 100 & 82 \\
all\_objects             &  6 & 100 & 100 \\
all\_classes             &  4 & 100 & 100 \\
object\_ratio            &  4 & 100 & 75 \\
\bottomrule
\end{tabular}
\end{minipage}

\vspace{6pt}
\centering\textbf{Overall Accuracy: 95.2\% (Gemma) / 74.4\% (Regex)}
\end{table}
\end{document}